\documentclass[letterpaper]{article} 
\usepackage{aaai24}  
\usepackage{times}  
\usepackage{helvet}  
\usepackage{courier}  
\usepackage[hyphens]{url}  
\usepackage{graphicx} 
\usepackage{natbib}  
\usepackage{caption} 
\usepackage{algorithm}
\usepackage{algorithm}
\usepackage{algorithmicx}
\usepackage[noend]{algpseudocode}
\usepackage{cleveref}
\usepackage{multirow}
\usepackage{array}
\usepackage{subcaption}

\usepackage{newfloat}
\usepackage{listings}
\DeclareCaptionStyle{ruled}{labelfont=normalfont,labelsep=colon,strut=off} 
\floatstyle{ruled}
\newfloat{listing}{tb}{lst}{}
\floatname{listing}{Listing}
\usepackage[dvipsnames]{xcolor}

\algdef{SE}[SWITCH]{Switch}{EndSwitch}[1]{\algorithmicswitch\ #1\ \algorithmicdo}{\algorithmicend\ \algorithmicswitch}%
\algdef{SE}[CASE]{Case}{EndCase}[1]{\algorithmiccase\ #1}{\algorithmicend\ \algorithmiccase}%
\algtext*{EndSwitch}%
\algtext*{EndCase}%
\algrenewcommand\textproc{}
\title{ITA-ECBS: A Bounded-Suboptimal Algorithm\\for the Combined Target-Assignment and Path-Finding Problem}
\author{
    Yimin Tang\textsuperscript{\rm 1}, 
    Sven Koenig\textsuperscript{\rm 1}, 
    Jiaoyang Li\textsuperscript{\rm 2}
}
\affiliations{

    \textsuperscript{\rm 1}University of Southern California, USA\\
    \textsuperscript{\rm 2}Carnegie Mellon University, USA\\
    \{yimint,skoenig\}@usc.edu, jiaoyangli@cmu.edu

}

\usepackage{bibentry}

\begin{document}

\maketitle

\begin{abstract}
Multi-Agent Path Finding (MAPF), i.e., finding collision-free paths for multiple robots, plays a critical role in many applications. Sometimes, assigning a target to each agent also presents a challenge. The Combined Target-Assignment and Path-Finding (TAPF) problem, a variant of MAPF, requires one to simultaneously assign targets to agents and plan collision-free paths for agents. Several algorithms, including CBM, CBS-TA, and ITA-CBS, optimally solve the TAPF problem, with ITA-CBS being the leading algorithm for minimizing flowtime.
However, the only existing bounded-suboptimal algorithm ECBS-TA is derived from CBS-TA rather than ITA-CBS. So, it faces the same issues as CBS-TA, such as searching through multiple constraint trees and spending too much time on finding the next-best target assignment. We introduce ITA-ECBS, the first bounded-suboptimal variant of ITA-CBS. Transforming ITA-CBS to its bounded-suboptimal variant is challenging because different constraint tree nodes can have different assignments of targets to agents. ITA-ECBS uses focal search to achieve efficiency and determines target assignments based on a new lower bound matrix. We show that it runs faster than ECBS-TA in 87.42\% of 54,033 test cases.
\end{abstract}

\section{Introduction}

The Multi-Agent Path Finding (MAPF) problem requires planning collision-free paths for multiple agents from their respective start locations to pre-assigned target locations in a known environment while minimizing a given cost function. Many algorithms have been developed to solve this problem optimally, such as Conflict-Based Search (CBS)~\cite{sharon2015conflict}, \(M^*\)~\cite{wagner2011m}, and Improved CBS (ICBS)~\cite{boyarski2015icbs}.
Solving the MAPF problem optimally is known to be NP-hard~\cite{yu2013structure}, so optimal MAPF solvers face challenges in scalability and efficiency.
In contrast, suboptimal MAPF solvers, such as Prioritized Planning (PP)~\cite{erdmann1987multiple,silver2005cooperative}, PBS~\cite{ma2019searching} and their variants~\cite{chan2023greedy,li2022mapf}, exhibit better scalability and efficiency. However, these algorithms lack theoretical guarantees for the quality of their solutions for a given cost function. Bounded-suboptimal MAPF solvers trade off between efficiency and solution quality. Enhanced CBS (ECBS)~\cite{barer2014suboptimal}, EECBS~\cite{li2021eecbs}, and LaCAM~\cite{okumura2023lacam} guarantee to find collision-free solutions whose costs are at most a user-defined suboptimality factor away from the optimal cost. 

Sometimes, assigning a target to
each agent also presents a challenge. This paper explores a variant of the MAPF problem, the Combined Target-Assignment and Path-Finding (TAPF) problem ~\cite{cbm2016,honig2018conflict}. Inspired by warehouse automation, where robots deliver shelves to packing stations and can initially select which shelf to retrieve ~\cite{wurman2008coordinating}, TAPF assigns each agent a target (location) from a set of possible targets. Subsequently, it finds collision-free paths for all agents to minimize a given cost function. TAPF is a more general problem and becomes MAPF if the size of the target set is restricted to one per agent. So, TAPF inherits the NP-hard complexity of MAPF.

Several algorithms have been proposed to solve the TAPF problem optimally, including CBM~\cite{cbm2016}, CBS-TA~\cite{honig2018conflict}, and ITA-CBS~\cite{tang2023solving}, with ITA-CBS being the state-of-the-art optimal algorithm for flowtime (i.e., the sum of all path costs). However, they all face scalability issues as optimal TAPF solvers. 
To the best of our knowledge, ECBS-TA~\cite{honig2018conflict} is the only existing bounded-suboptimal TAPF solver. It directly applies the ECBS algorithm to CBS-TA and can find collision-free (valid) solutions more quickly than CBS-TA. However, since ECBS-TA is based on CBS-TA, ECBS-TA encounters efficiency problems due to the same two issues as CBS-TA: (1) ECBS-TA maintains multiple Constraint Trees (CT) and explores each sequentially, leading to many CT nodes. (2) It involves solving a K-best target assignment problem~\cite{chegireddy1987algorithms}, which is often time-consuming. To address these issues, we have developed a bounded-suboptimal algorithm inspired by the single CT structure of ITA-CBS, aiming to avoid the computational bottlenecks of ECBS-TA.

While ITA-CBS is a CBS-like algorithm with a single CT, developing a bounded-suboptimal algorithm from ITA-CBS is not straightforward since the target assignment (TA) solution, an arrangement that specifies a target for each agent, varies at each CT node. 
Simply applying ECBS to ITA-CBS can lead to the returned valid solution not being bounded by a suboptimal factor. By incorporating an additional Lower Bound (LB) matrix and deriving the TA solution from it, we develop Incremental Target Assignment with Enhanced CBS (ITA-ECBS), a bounded-suboptimal variant of ITA-CBS with flowtime. It can avoid producing unbounded valid solutions, a risk present when directly applying ECBS to ITA-CBS. Furthermore, it uses the shortest path costs as LB values, thereby accelerating path searching algorithm. Our experimental results show that ITA-ECBS runs faster than the baseline algorithm ECBS-TA in 87.42\% of 54,033 test cases with 8 different suboptimality factors.

\section{Problem Definition}

The Combined Target-Assignment and Path-Finding (TAPF) problem is defined as follows: Let $I=\{1,2,\cdots,N\}$ denote a set of $N$ agents. $G = (V,E)$ represents an undirected graph, where each vertex $v \in V$ represents a possible location of an agent in the workspace, and each edge $e \in E$ is a unit-cost edge between two vertices that moves an agent from one vertex to the other. Self-loop edges are allowed, which represent ``wait-in-place'' actions.
Each agent $i\in I$ has a start location $s_i \in V$.
Let $\mathcal{G}=\{g_1, g_2, \cdots, g_M\} \subseteq V$ denote a set of $M$ targets ($M\geq N$).
Let $A$ denote a binary $N \times M$ \emph{target matrix}, where each entry $A[i][j]$ (the $i$-th row and $j$-th column in $A$) is one if agent $i$ is eligible to be assigned to target $g_j$ and zero otherwise. We refer to the set of targets $\{g_j \in \mathcal{G} | A[i][j]=1\}$ as the \emph{target set} for agent $i$. 
Our task is to assign each agent $i$ a target $g_j$ from its target set and plan corresponding collision-free paths for all agents. We cannot assign an agent without specifying a target.

Each action of agents, either waiting in place or moving to an adjacent vertex, takes one time unit.
Let $v^i_t \in V$ be the location of agent $i$ at timestep $t$.
Let $\pi_i=[v_0^{i}, v_1^{i}, ..., v_{T^{i}}^{i}]$ denote a path of agent $i$ from its start location $v_0^{i}$ to its target $v_{T^{i}}^{i}$. 
We assume that agents rest at their targets after completing their paths, i.e., $v_t^i = v_{T^i}^i, \forall t > T^i$.
The cost of agent $i$'s path is $T^i$. We refer to the path with the minimum cost as the shortest path. We consider two types of agent-agent collisions.
The first type is a \emph{vertex collision}, where two agents $i$ and $j$ occupy the same vertex at the same timestep. 
The second type is an \emph{edge collision}, where two agents move in opposite directions along the same edge.
We use $(i, j, t)$ to denote a vertex collision between agents $i$ and $j$ at timestep $t$ or a edge collision between agents $i$ and $j$ at timestep $t$ to $t+1$.
The requirement of being collision-free implies the targets assigned to the agents must be distinct from each other.

The objective of the TAPF problem is to find a set of paths $\{\pi_i | i\in I\}$ for all agents such that, for each agent $i$:
\begin{enumerate}
\item Agent $i$ starts from its start location (i.e., $v_0^{i} = s_i$);
\item Agent $i$ stops at a target $g_j$ in its target set (i.e., $v_{t}^{i} = g_j, \forall t \ge T^{i}$ and $A[i][j] = 1$);
\item Every pair of adjacent vertices on path $\pi_i$ is connected by an edge (i.e., $(v_{t}^{i}, v_{t+1}^{i}) \in E, \forall 0 \leq t \le T^i$); and
\item $\{\pi_i | i\in I\}$ is collision-free and minimize the \emph{flowtime} $\sum_{i=1}^{N}T^{i}$.
\end{enumerate}

\section{Related Work}

\subsection{Focal Search}
 
Focal search~\cite{pearl1982studies,cohen2018anytime} is bounded-suboptimal search. Given a user-defined suboptimality factor $w \geq 1$, it is guaranteed to find a solution with a cost at most $w \cdot c^{opt}$, where $c^{opt}$ is the cost of an optimal solution. Focal search has two queues: OPEN and FOCAL. OPEN stores all candidates that need to be searched and sort each candidate $n$ by $f(n)=g(n)+h(n)$, where \(g(n)\) and \(h(n)\) are the cost and an admissible heuristic value of candidate $n$, respectively, which are identical to the $g$ and $h$ values in \(A^*\) search. FOCAL includes all candidates $n$ satisfying $f(n) \leq w \cdot f_{front}$, where $f_{front}$ is the minimum $f$ value in OPEN. FOCAL sorts candidates by another heuristic function $d(n)$ which could be any function defined by users.\footnote{Since OPEN and FOCAL are for sorting purposes and FOCAL is a subset of OPEN, in implementation, we store candidate pointers in them. If one candidate appears in both queues, only one copy exists and two pointers point to this candidate copy.} Focal search searches candidates in order of FOCAL—the FOCAL aids in quickly identifying a solution through its heuristic function. When we find a solution with cost $c^{val}$, we call current $f_{front}$ as this solution's lower bound (LB) because of $f_{front} \leq c^{opt}$ and $f_{front} \leq c^{val} \leq wf_{front}$. Therefore, focal search outputs two key pieces of information: an LB value $c^g$ and a solution with cost $c$. If there is no solution, we set both $c^g$ and $c$ to $\infty$. 

\subsection{Multi-Agent Path Finding (MAPF)}

MAPF has a long history~\cite{silver2005cooperative}, and many algorithms have been developed to solve it or its variants. The problem is to find collision-free paths for multiple agents from their start locations to pre-assigned targets while minimizing a given cost function. Decoupled algorithms~\cite{silver2005cooperative,luna2011push,wang2008fast} independently plan a path for each agent and then combine all paths to one solution. Coupled algorithms~\cite{standley2010finding,standley2011complete} plan for all agents together. There also exist dynamically-coupled algorithms~\cite{sharon2015conflict,wagner2015subdimensional} that independently plan each agent and re-plan multiple agents together when needed to resolve their collisions.
Among them, Conflict-Based Search (CBS)~\cite{sharon2015conflict} is a popular centralized optimal MAPF algorithm. Some bounded-suboptimal algorithms are based on it, such as ECBS~\cite{barer2014suboptimal} and EECBS~\cite{li2021eecbs}. 

\paragraph{CBS}
Conflict-Based Search (CBS) is an optimal two-level search algorithm. 
Its low level plans the shortest paths for agents from their start locations to targets, while its high level searches a binary Constraint Tree (CT). 
Each CT node \(H = (c, \Omega, \pi)\) includes a constraint set \(\Omega\), a solution \(\pi\), which is a set of shortest paths satisfying \(\Omega\) for all agents, and a cost \(c\), which is the flowtime of \(\pi\). When a solution \( \pi \) or a path does not include any agent actions or positions that are restricted by a \(\Omega\), we say this solution or path satisfies the \(\Omega\). As long as \(\pi\) satisfies \(\Omega\), \(pi\) could have collisions. We call a CT node solution \(\pi\) a valid solution when it is collision-free.
When expanding a node \(H\), CBS selects the first collision in \(H.\pi\), even when multiple collisions occur in \(H.\pi\), and formulates two constraints, each prohibiting one agent from occupying the colliding location or executing its intended original action at the colliding timestep. We have two types of constraints: vertex constraint $(i, v, t)$ that prohibits agent $i$ from occupying vertex $v$ at timestep $t$ and edge constraint $(i, u, v, t)$ that prohibits agent $i$ from going from vertex $u$ to vertex $v$ at timestep $t$. Then CBS generates two successor nodes identical to \(H\) and adds each of the two constraints to the constraint set of the two respective successor nodes. After adding a new constraint, each node should re-plan the path that does not satisfy this constraint. By maintaining a priority queue OPEN based on each node's cost, CBS repeats this process until expanding a node that has no collisions, in which case, its solution is an optimal valid solution. CBS is optimal for the flowtime minimization~\cite{sharon2015conflict}.

\paragraph{ECBS} 
Enhanced CBS (ECBS)~\cite{barer2014suboptimal} is based on CBS and uses focal search with the same suboptimality factor $w$ in both two-level searches. 
In its low-level search, the focal search returns an LB value \(c^g_i\) and a valid path \(\pi_i\) with cost \(c_i\) for agent $i$ from its start location to target. The path and value satisfy: \(c^g_i \leq c_i^{opt} \leq c_i \leq w \cdot c^g_i\), where \(c_i^{opt}\) is the cost of agent \(i\)'s shortest path satisfying \(\Omega\), and \(w\) is the suboptimality factor. In its high-level search, comparing to CBS, each CT node \(H = (c, \Omega, \pi, L, c_L)\) in ECBS has an additional cost array \(L\), which stores all LB values $c^g_i$ for all paths in \(\pi\), and cost \(c_L\) which is the sum of $c^g_i$ in \(L\). The high-level search of ECBS maintains two priority queues: FOCAL and OPEN. OPEN stores all CT nodes sorted in ascending order of \(c_L\). Let the front CT node in OPEN be  \(H_{front}\), ECBS adds all CT nodes \(H\) in OPEN that satisfy \(H.c \leq w \cdot H_{front}.c_L\) into FOCAL. FOCAL is sorted in ascending order of a user-defined heuristic function \(d(H)\). ECBS guarantees that its returned solution \(H^{sol}.\pi\) satisfies \(H^{sol}.c \leq w \cdot c^{opt}\), where \(c^{opt}\) is the cost of the optimal valid solution.

\subsection{Combined Target-Assignment and Path-Finding (TAPF)}

The Combined Target-Assignment and Path-Finding (TAPF) problem is a combination of the MAPF problem and the target-assignment problem. While MAPF has a pre-defined target for each agent, TAPF involves simultaneously assigning targets to agents and finding collision-free paths for them. There are several TAPF algorithms such as CBM~\cite{cbm2016}, CBS-TA~\cite{honig2018conflict}, ECBS-TA~\cite{honig2018conflict}, and ITA-CBS~\cite{tang2023solving}. CBM combines CBS with maxflow algorithms to optimally minimize the makespan \(\max_{i \in I}\{T^i\}\). However, CBM works only for makespan, while other algorithms also work for flowtime. CBS-TA tries different Target Assignments (TA) for agents and then seeks the optimal valid solution across multiple CTs, one CT for each TA (forming a forest). Targets are assigned to agents by Hungarian algorithm, that minimizes the sum of costs in a given cost matrix. ECBS-TA, a bounded-suboptimal version of CBS-TA, incorporates focal search~\cite{pearl1982studies,barer2014suboptimal}. 
Both CBS-TA and ECBS-TA require a substantial amount of time to determine the next-best TA as they lazily traverse each CT. 
In contrast to CBS-TA and ECBS-TA, ITA-CBS searches for optimal valid solutions within a single CT and uses the dynamic Hungarian algorithm~\cite{mills2007dynamic}. 
However, scalability remains a challenge for ITA-CBS due to its optimality. We propose ITA-ECBS to overcome this issue.

\paragraph{CBS-TA and ECBS-TA}
CBS-TA~\cite{honig2018conflict}, inspiring many extensions~\cite{ren23cbss,zhong2022optimal,chen2021integrated,okumura2023solving}, operates on the following principle: a fixed TA transforms a TAPF instance to a MAPF instance, and CBS can solve each MAPF instance with one CT. CBS-TA efficiently explores all nodes of the different CTs (CT forest) by enumerating every TA solution. In CBS-TA, TA solutions are derived from an $N \times M$  cost matrix $M_c$, which records the path costs from agents' start locations to targets, without considering any constraints.
Each CT node \(H = (c, \Omega, \pi, \pi_{ta}, r)\) of CBS-TA has two extra fields compared to a node of CBS: a TA solution \(\pi_{ta}\) that assigns each agent a unique target 
and a root flag \(r\) signifying if \(H\) is the root of a CT. 
CBS-TA maintains a priority queue OPEN to store nodes from the CT forest and lazily generates roots of new CTs with different TA solutions.
CBS-TA first generates one CT root with the optimal TA which has the minimum cost based on $M_c$.
It does not need to generate a new CT until all CT nodes in the queue are larger than the cost of a new CT root because the cost of a CT root is no larger than the cost of any child node in the CT. 
Consequently, it generates a new root with the next-best TA solution only when the CT root in OPEN has been expanded.
Based on the K-best task-assignment algorithm~\cite{chegireddy1987algorithms} and the Successive Shortest Path (SSP) algorithm~\cite{engquist1982successive}, CBS-TA finds the next-best TA solution with complexity $O(N^3M)$. Using the ECBS algorithm to search each CT transforms CBS-TA to the bounded-suboptimal algorithm ECBS-TA~\cite{honig2018conflict}.

\paragraph{ITA-CBS}
Since many CT nodes in different CTs of CBS-TA have identical constraint sets, which leads CBS-TA to repeat searches, ITA-CBS is a CBS-like algorithm that uses only one CT to search for the optimal valid solution. Each CT node \(H = (c, \Omega, \pi, \pi_{ta}, M_c)\) in ITA-CBS has two extra fields compared to a node of CBS: a TA solution \(\pi_{ta}\) and an $N\times M$ cost matrix $M_c$. Each entry $M_c[i][j]$ is the minimum cost of paths from $s_i$ to $g_j$ satisfying \(\Omega\). $M_c[i][j]=\infty$ if $A[i][j]=0$ (i.e., target $g_{j}$ is not included in the target set of agent $i$) or there is no path satisfying \(\Omega\). In implementation, ITA-CBS stores costs in \(M_c\) together with their corresponding paths, so that it can construct $\pi$ directly from $M_c$ after obtaining \(\pi_{ta}\). In this paper, when we mention $M_c$, it could represent costs or the corresponding paths depending on context. $\pi_{ta}$ is the optimal TA solution of $M_c$, $\pi$ is the set of paths selected from $M_c$ based on $\pi_{ta}$, and $c$ is the flowtime of $\pi$. 

During node expansion, ITA-CBS retrieves a node from OPEN and checks if its $\pi$ is collision-free. If so, this $\pi$ is an optimal valid solution, and ITA-CBS terminates. Otherwise, like CBS, ITA-CBS generates two child CT nodes and adds new constraints to their \(\Omega\)s. 
When creating a CT node \(H\), ITA-CBS first re-plans all paths in $M_c$ that do not satisfy \(H.\Omega\). Then, it obtains a new \(H.\pi_{ta}\) from \(H.M_c\). Based on \(H.\pi_{ta}\), ITA-CBS obtains a new \(H.\pi\) from \(H.M_c\) and then inserts $H$ into OPEN. In summary, the order of modification of the variables is $\Omega \rightarrow M_c \rightarrow \pi_{ta} \rightarrow \pi \rightarrow$ children's $\Omega$. 
Since each CT node has only one new constraint compared to its parent and at most one row in $M_c$ is changed, ITA-CBS uses dynamic Hungarian algorithm~\cite{mills2007dynamic} with complexity \(O(NM)\) to obtain a new TA, which largely reduces the runtime of TA compared to using Hungarian algorithm. ITA-CBS is faster than CBS-TA in experiments. However, its scalability is still limited since it is an optimal algorithm for TAPF.

\begin{figure}[t!]
\centering
\includegraphics[width=0.4\textwidth]{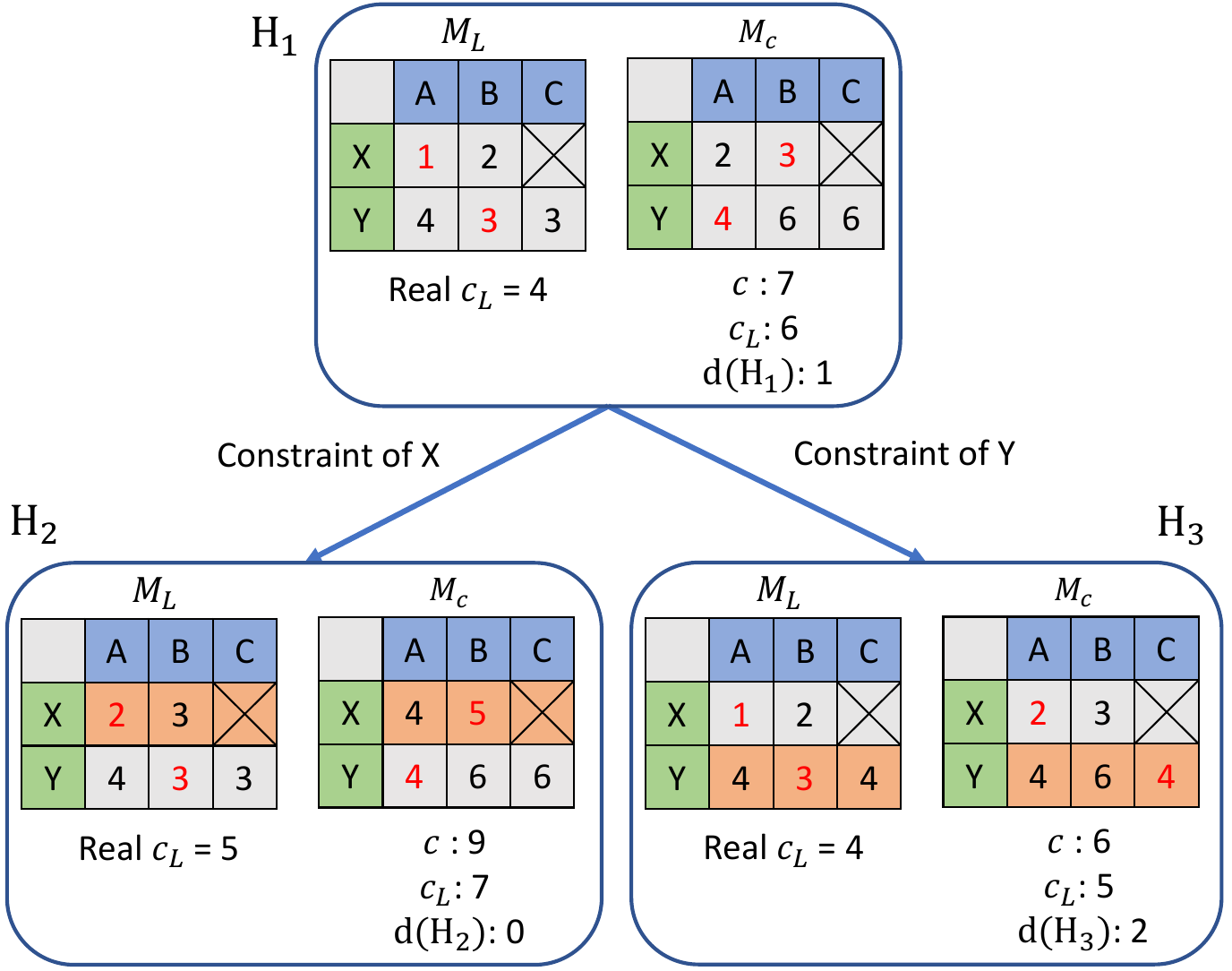} 
\caption{This figure shows the unbounded problem if we directly use ECBS in ITA-CBS. We have 2 agents \(\{X,Y\}\) and target sets $X$:\(\{A, B\}\) and $Y$:\(\{A,B,C\}\). Red numbers represent the matrix's optimal TA solution. Orange cells represent the row update since the new constraint is related to only one agent. The suboptimality factor $w$ is $2$. ITA-CBS utilizes \(M_c\) to obtain the 
optimal TA solution. Here, $\pi_{ta}$ is the TA solution of $M_c$. \(c\) represents the flowtime based on $\pi_{ta}$. \(c_L\) is the sum of LB values selected from $M_L$ based on $\pi_{ta}$. Real \(c_L\) is the cost of the optimal TA solution of \(M_L\). \(d(H)\) represents the user-defined heuristic function used in FOCAL.
}
\label{fig:ta_on_LB}
\end{figure}

\begin{algorithm}[th!]
\small
\caption{ITA-ECBS-v0 and ITA-ECBS}
\label{alg:ITA-ECBS1}
\textbf{Input}: Graph \(G\), start locations $\{s_i\}$, target locations $\{g_i\}$, target matrix $A$, suboptimality factor $w$, algorithm type ALGO\_TYPE\\
\textbf{Output}: A valid TAPF solution within the suboptimality factor $w$ 

\begin{algorithmic}[1] 
\State \(H_{0}\) = new CTnode()
\State \(H_{0}.\Omega\) = $\emptyset$
\For{\textbf{each} $(i,j) \in \{1, \cdots, N\} \times \{1, \cdots, M\}$}
\State \(H_{0}.M_c\)[$i$][$j$] = \(H_{0}.M_L\)[$i$][$j$] = $\infty$
\If {$A[i][j]=1$}
    \State \(H_{0}.M_L\)[$i$][$j$],\(H_{0}.M_c\)[$i$][$j$]=lowLevelSearch($G, s_i, g_j, H_{0}, w$)
\EndIf
\EndFor
\State \(H_{0}.\pi_{ta}\) = optimalTargetAssignment(\(H_{0}.M_L\))
\State \(H_{0}.c, H_{0}.\pi\) = getPlan(\(H_{0}.\pi_{ta}\), \(H_{0}.M_c\))
\State \(H_0.c_L\) = getLowerBound(\(H_0.\pi_{ta}\), \(H_0.M_L\))
\State FOCAL = OPEN = PriorityQueue()
\State Calculate \(d(H_0)\) and insert \(H_{0}\) into OPEN
\While{OPEN not empty}
\State \(H_{front}\) = OPEN.front() 
\State FOCAL = FOCAL $\cup$ \{H $\in$ OPEN $\mid$ $H.c \leq w \cdot H_{front}.c_L$\}
\State \(H_{cur}\) = FOCAL.front(); FOCAL.pop()
\State Delete \(H_{cur}\) from OPEN
\If {\(H_{cur}.\pi\) has no collision}
    \State \textbf{return} \(H_{cur}.\pi\)
\EndIf
\State ($i, j, t$) = getFirstCollision($H_{cur}.\pi$)
\For{\textbf{each} agent $k$ in ($i,j$)}
    \State $Q$ = a copy of \(H_{cur}\)
    \If{($i, j, t$) is a vertex collision}
        \State \(Q.\Omega\) = \(Q.\Omega\) $\cup$ ($k$, $v^k_{t}$, $t$) // vertex constraint
    \Else 
        \State \(Q.\Omega\) = \(Q.\Omega\) $\cup$ ($k$, $v^k_{t-1}$, $v^k_{t}$, $t$) // edge constraint
    \EndIf
    \For{\textbf{each} $x$ with $A[k][x]=1$}
        \State \(Q.M_L\)[$k$][$x$],\(Q.M_c\)[$k$][$x$]=lowLevelSearch($G, s_k, g_x, Q, w$)
    \EndFor
    \State \(Q.\pi_{ta}\) = optimalTargetAssignment(\(Q.M_L\))
    \State \(Q.c, Q.\pi\) = getPlan(\(Q.\pi_{ta}\), \(Q.M_c\))
    \State \(Q.c_L\) = getLowerBound(\(Q.\pi_{ta}\), \(Q.M_L\))
    \If{Q.c $\textless$ $\infty$}
        \State Calculate \(d(Q)\) and insert $Q$ into OPEN
    \EndIf
\EndFor
\EndWhile
\State \textbf{return} No valid solution

\Function{lowLevelSearch}{$G$, $s_k$, $g_x$, $Q, w$}
\If {ALGO\_TYPE = ITA-ECBS-v0}
    \State \(c^g\), \(c\) =  focalSearch($G$, $s_k$, $g_x$, \(Q.\Omega\), $Q.M_c$, $w$)
\EndIf
\If {ALGO\_TYPE = ITA-ECBS}
    \State \(c^g\) = shortestPathSearch($G$, $s_k$, $g_x$, \(Q.\Omega\))
    \State \(c\) =  searchWithLB($G$, $s_k$, $g_x$, \(Q.\Omega\), $Q.M_c$, $w$, \(c^g\))
\EndIf
\State \textbf{return} \(c^g\), \(c\)
\EndFunction
\end{algorithmic}
\end{algorithm}

\begin{figure*}[htb!]
\centering
     \begin{subfigure}[b]{0.32\textwidth}
         \centering
         \includegraphics[width=\textwidth]{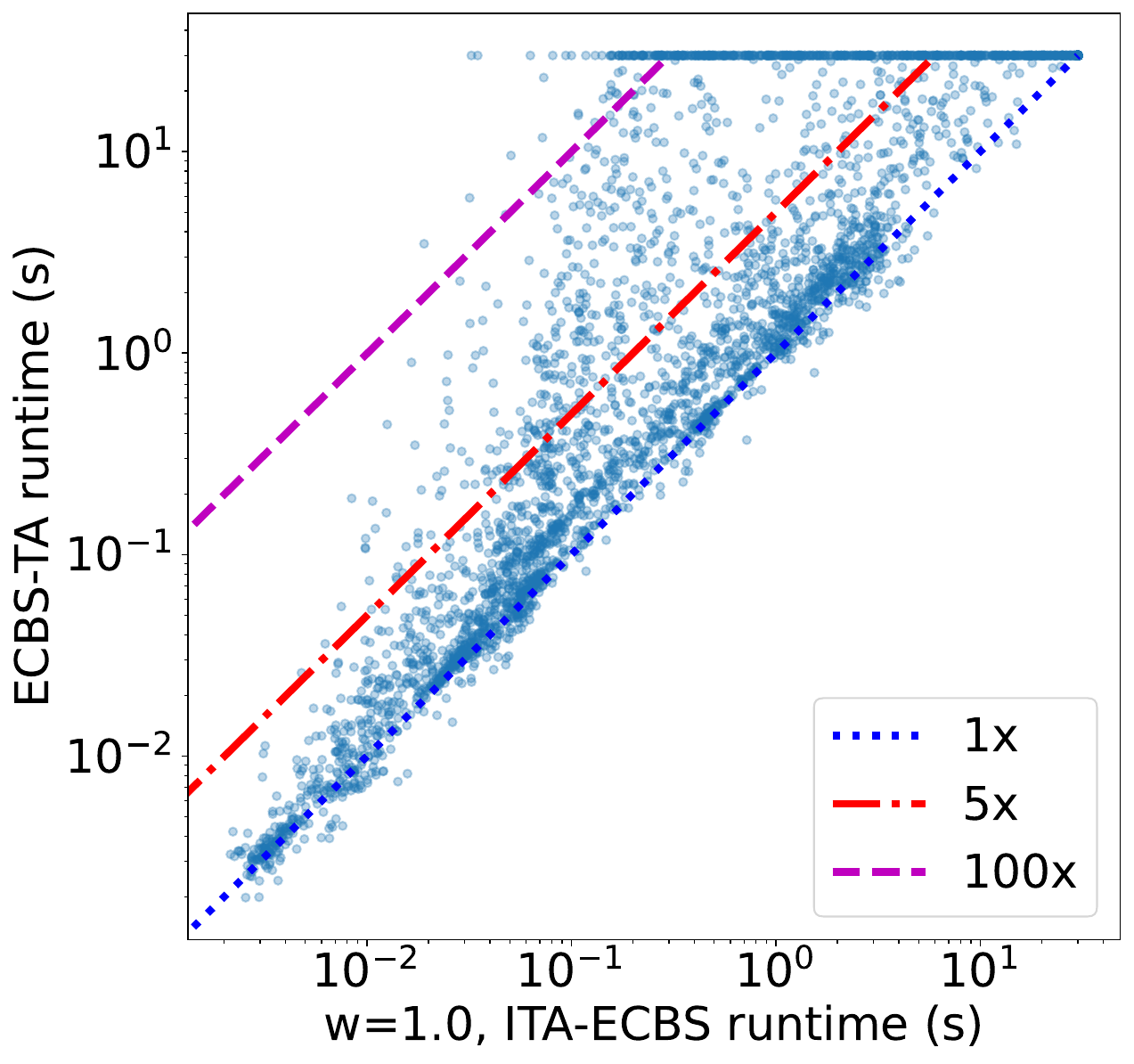}
         \label{fig:w100}
     \end{subfigure}
     \begin{subfigure}[b]{0.32\textwidth}
         \centering
         \includegraphics[width=\textwidth]{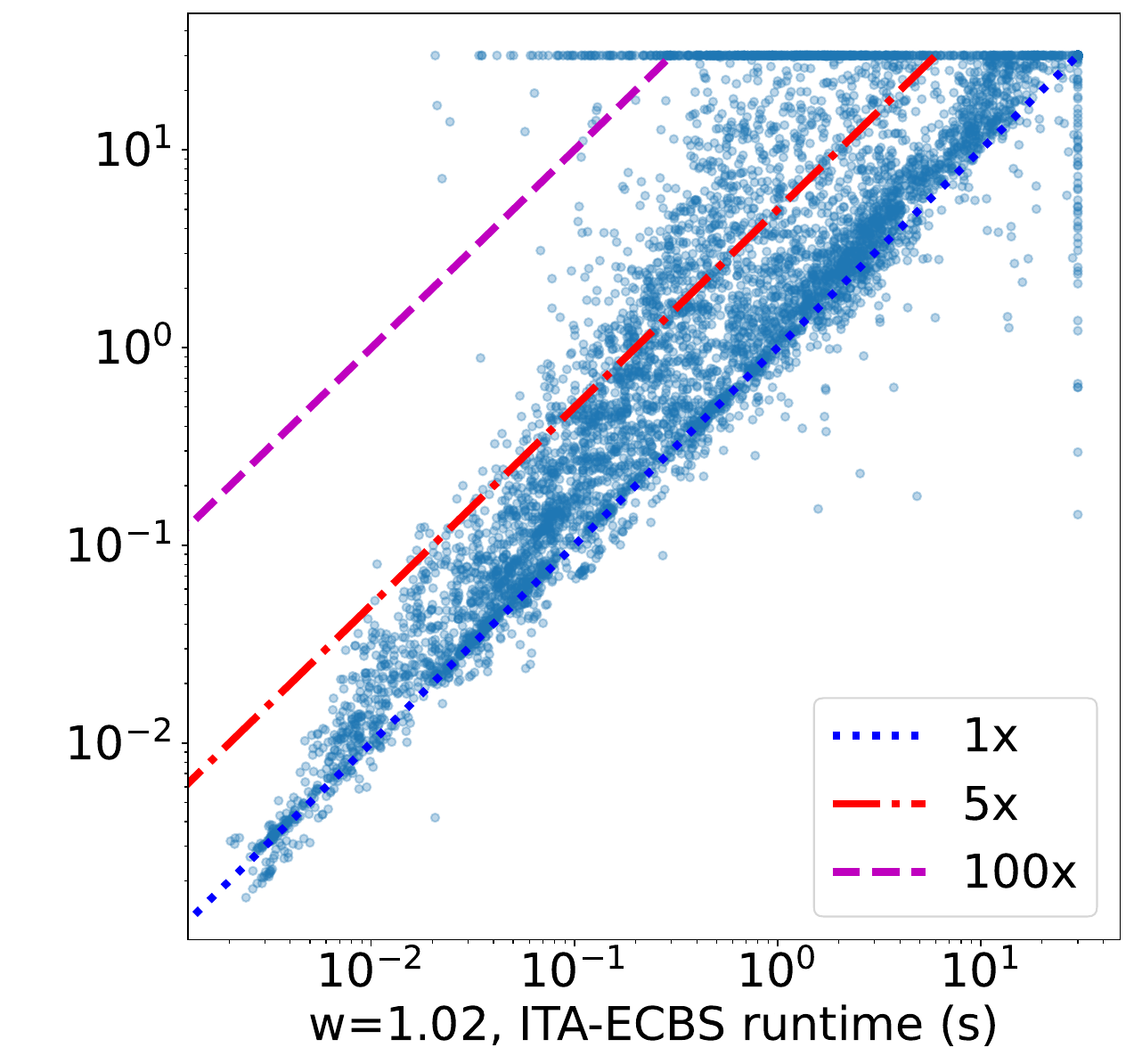}
         \label{fig:w102}
     \end{subfigure}
     \begin{subfigure}[b]{0.32\textwidth}
         \centering
         \includegraphics[width=\textwidth]{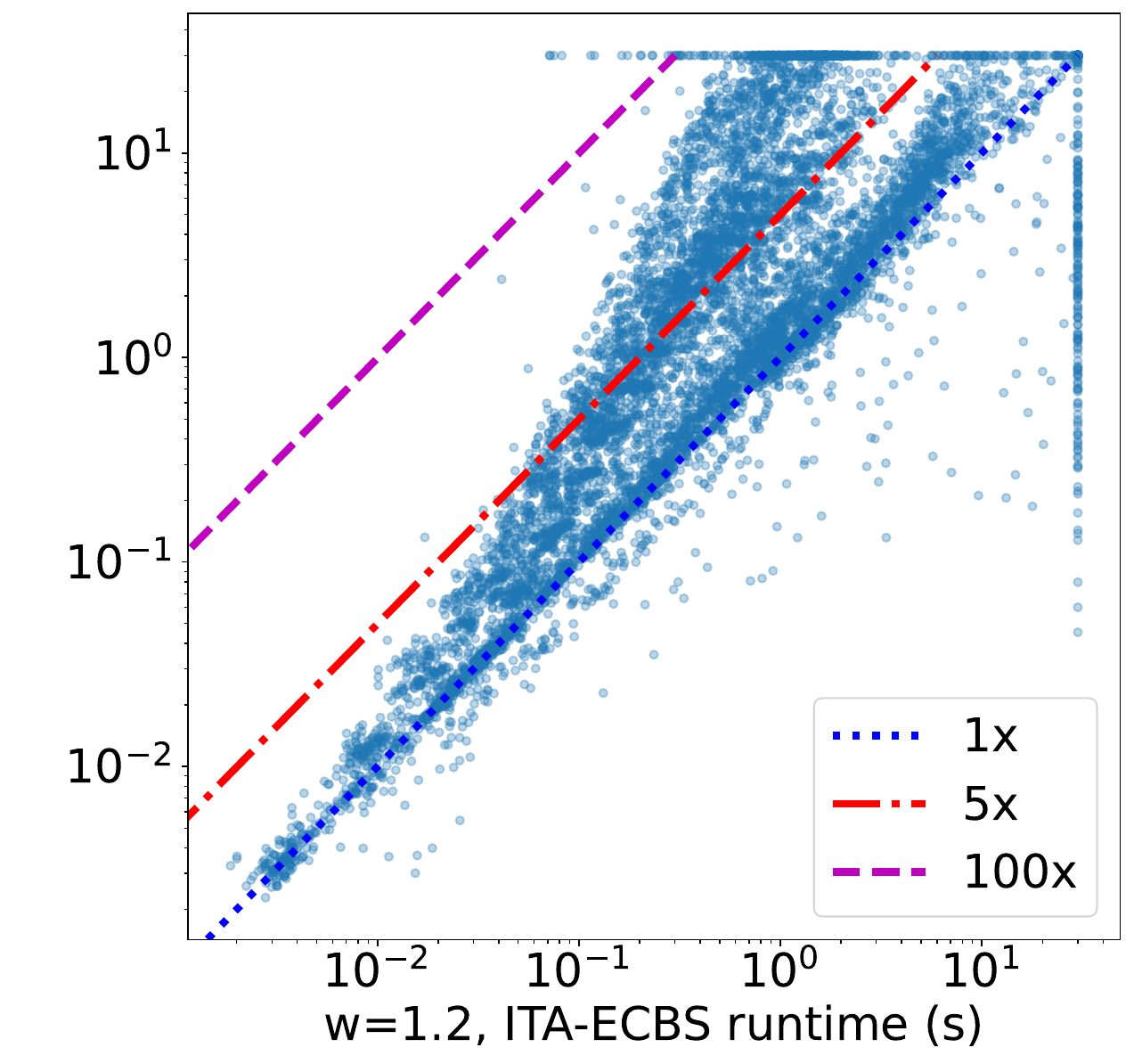}
         \label{fig:w120}
     \end{subfigure}
    \caption{Each subfigure contains 9,600 $(8\cdot 15\cdot 4\cdot 20)$ distinct test cases. We have chosen three suboptimality factors $w$ \{1.00, 1.02, 1.20\} to find optimal and bounded-suboptimal valid solutions with small and large suboptimality factors.
    }
    \label{fig:g3}
\end{figure*}

\begin{figure}[t!]
\centering
\includegraphics[width=0.48\textwidth]{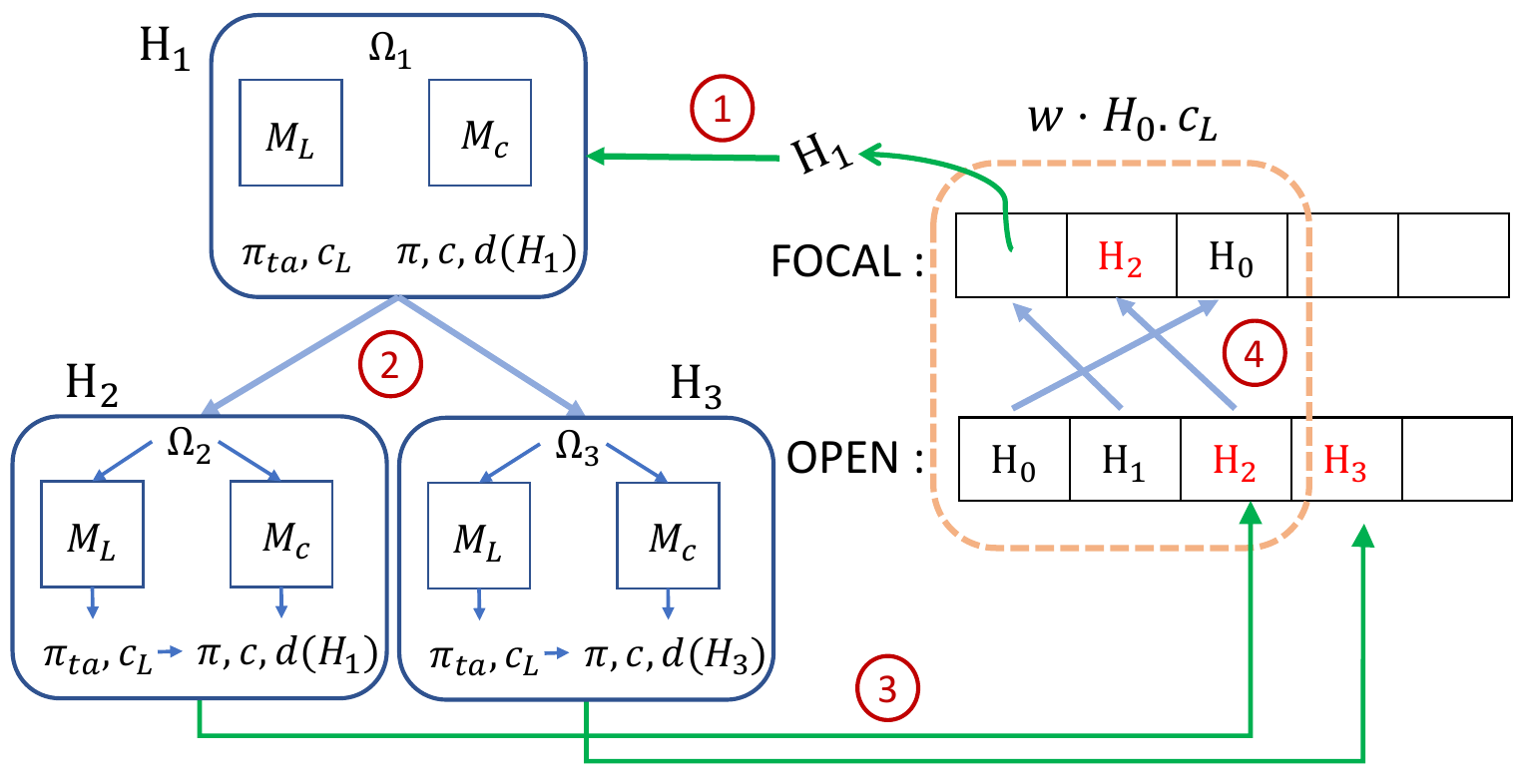} 
\caption{ITA-ECBS overview: There are two CT nodes \{\(H_0, H_1\)\} in OPEN. (1) Although \(H_0\) has a lower \(c_L\) value and precedes \(H_1\) in OPEN, the heuristic function \(d(H)\) may result in \(H_1\) being selected for expansion. (2) We verify whether \(H_1.\pi\) is collision-free. If not, we generate two child CT nodes with new constraint sets \(\Omega_2\) and \(\Omega_3\) and then use focal search to obtain new \(M_L\) and \(M_c\) for each node. A new \(M_L\) leads to a new \(\pi_{ta}\) and \(c_L\). The updated \(\pi_{ta}\) and \(M_c\) give us a bounded-suboptimal solution \(\pi\) and the cost $c$. We then calculate $d(H)$. (3) We insert these two nodes into OPEN, indicated by the red values. (4) All CT nodes in OPEN with cost \(c \leq w \cdot H_0.c_L\) are added to FOCAL. \(H_2\) could be positioned ahead of \(H_0\) in FOCAL by the heuristic function \(d\).
}
\label{fig:algo_overall}
\end{figure}

\section{ITA-ECBS}

Adapting the optimal ITA-CBS algorithm to its bounded-suboptimal counterpart is challenging since the optimal TA often changes from CT node to CT node. In ITA-CBS, we derive \(\pi_{ta}\) from \(M_c\) and a solution \(\pi\) based on \(\pi_{ta}\). Directly applying focal search during low-level path search in ITA-CBS can yield two $N \times M$ matrices for each CT node $H$: an LB matrix \(M_L\), which stores all LB values $c^g$ returned from focal search, and a cost matrix $M_c$ satisfying $M_L[i][j] \leq M_c[i][j] \leq w\cdot M_L[i][j], i=1, ..., N, j=1, ..., M$. We can apply a target assignment algorithm to either one of these two matrices: 
\(\pi_{ta}\) obtained from \(M_L\) provides a lower bound on the cost of all TAPF solutions that meet \(H.\Omega\), whereas \(\pi_{ta}\) obtained from \(M_c\) has the minimum flowtime among all possible TA solutions in $M_c$. These two TA can differ. If we simply apply ECBS in ITA-CBS to make it bounded-suboptimal version, the returned valid solution \(\pi\), derived from \(\pi_{ta}\) of \(M_c\), may not be bounded-suboptimal. 

\Cref{fig:ta_on_LB} provides an example with suboptimality factor $w=2$. \(H_1\) is a parent CT node with $\pi_{ta}=\{X \rightarrow B, Y \rightarrow A\}$ based on \(H_1.M_c\). The sum of LB values selected from $H_1.M_L$ based on $\pi_{ta}$ is $c_L=6$, whereas the minimum LB sum $Real\ c_L$ is $4$ in \(M_L\). $H_1$ generates two child nodes \{$H_2,H_3$\}, assuming both child node solutions \(\pi\) are collision-free. Since $H_2.c \leq w \cdot \min(H_2.c_L, H_3.c_L)$, we add $H_2$ to FOCAL and the same applies to $H_3$. Assume $H_2$ has a lower heuristic value $d(H_2)$ and should be evaluated before $H_3$, leading to $H_2.\pi$ becoming the returned valid solution. The flowtime of $H_2.\pi$ is $9$. However, based on $Real\ c_L=4$ in $H_3$, it might be possible to find a child node of $H_3$ that has a valid solution with a flowtime $c^{opt}$ equal to $Real\ c_L=4$. In this case, $H_2.c \geq w\cdot c^{opt}$. This suggests that $H_2$'s solution may not be bounded by $w$, which we call the unbounded problem.

Focal search thus cannot turn ITA-CBS into a bounded-suboptimal algorithm. We need to obtain \(\pi_{ta}\) from $M_L$ rather than $M_c$. As shown in \Cref{fig:algo_overall} and \Cref{alg:ITA-ECBS1}, we propose two bounded-suboptimal TAPF algorithms: ITA-ECBS-v0 and its enhanced version ITA-ECBS. We first introduce ITA-ECBS-v0. Each CT node of ITA-ECBS-v0 is represented as \(H = (c, \Omega, \pi, \pi_{ta}, M_c, M_L, c_L)\). It has two extra fields compared to a node of ITA-CBS: an LB matrix $M_L$ and cost $c_L$ which is the sum of LB values selected from $M_L$ based on $\pi_{ta}$. The two matrices $M_L$ and $M_c$ can be generated concurrently because focal search returns an LB value and a path satisfying \(\Omega\) in one search. Each entry $M_L[i][j]$ represents the LB value of paths from $s_i$ to $g_j$, and $M_c[i][j]$ denotes the actual path and $M_c[i][j] \leq w\cdot M_L[i][j]$. If focal search has no solution for a path satisfying \(\Omega\) or $A[i][j]=0$, we set both $M_L[i][j]$ and $M_c[i][j]$ to $\infty$. A key distinction of ITA-ECBS-v0 from ITA-CBS is that $\pi_{ta}$ is the optimal TA solution obtained from $M_L$ rather than $M_c$. This change helps to avoid the unbounded problem. $\pi$ is the set of paths selected from $M_c$ based on $\pi_{ta}$. $c$ is the flowtime of $\pi$.

\begin{figure*}[t!]
\centering
\includegraphics[width=1\textwidth]{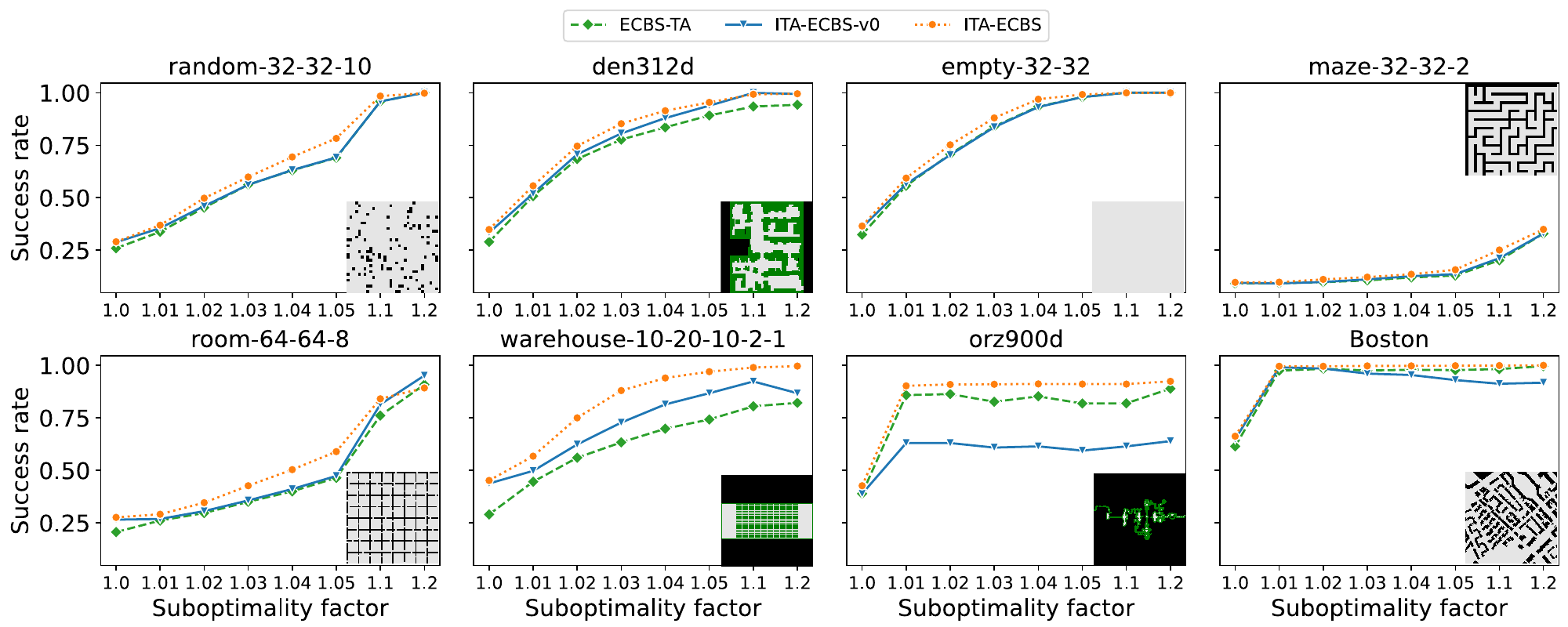}
\caption{The success rates of different algorithms as a function of the suboptimality factor.}
\label{fig:w_cr}

\includegraphics[width=1\textwidth]{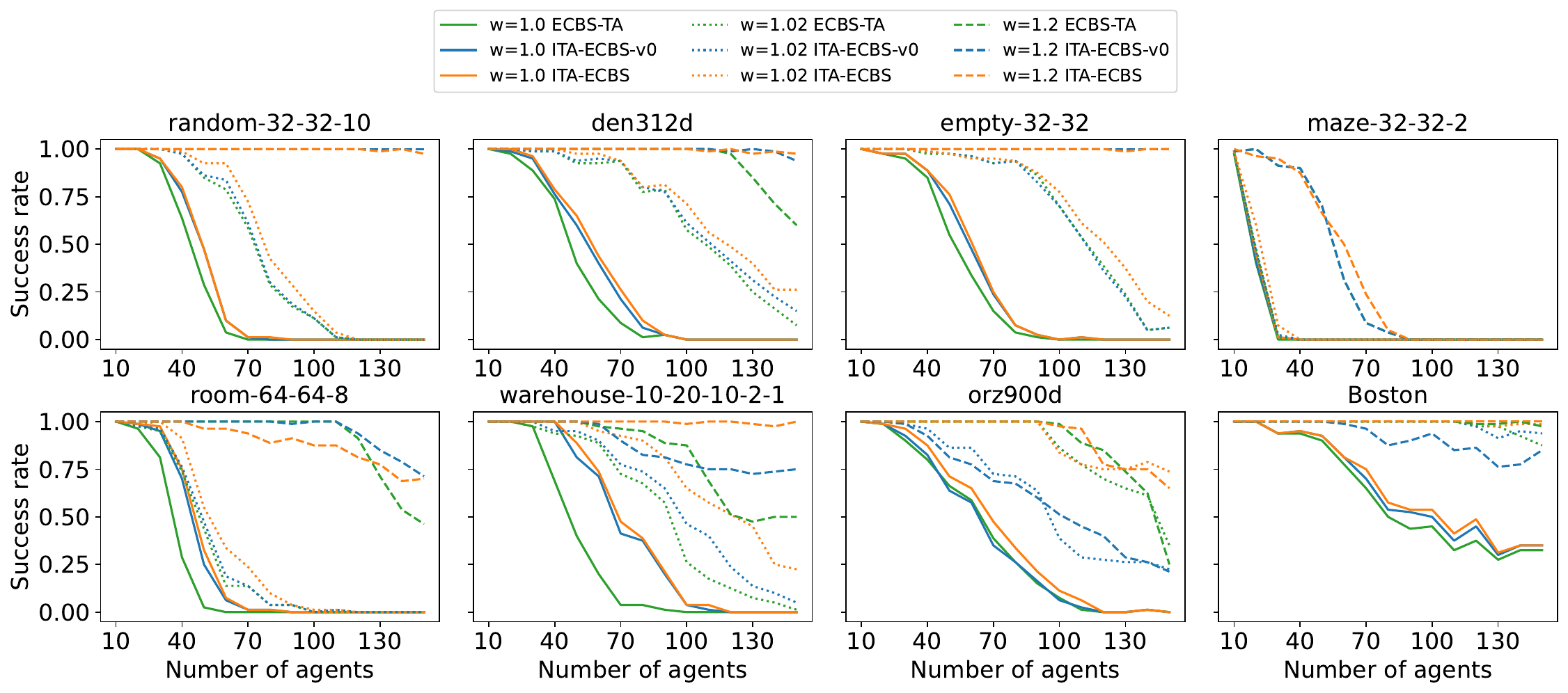}
\caption{The success rates of three algorithms with three selected suboptimality factors as a function of the number of agents.}
\label{fig:w_anum_cr}
\end{figure*}

\begin{table*}[t!]
\centering
\begin{tabular}{|c|c|m{1cm}|m{1cm}|m{1cm}|m{1cm}|m{1cm}|m{1cm}|m{1cm}|m{1cm}|m{1cm}|}
\hline
\multicolumn{2}{|c|}{\textbf{Average of Success Rate}} & \multicolumn{5}{c|}{\textbf{Number of Agents}} & \multicolumn{4}{c|}{\textbf{Percentage of Shared Targets}} \\ \hline
\textbf{$w$} & \textbf{algorithms} & \centering\textbf{30} & \centering\textbf{60} & \centering\textbf{90} & \centering\textbf{120} & \centering\textbf{150} & \centering\textbf{0\%} & \centering\textbf{30\%} & \centering\textbf{60\%} & \centering\arraybackslash\textbf{100\%} \\ \hline
\multirow{3}{*}{\textbf{1.01}} & ECBS\_TA & \centering 0.854 & \centering 0.582 & \centering 0.396 & \centering 0.240 & \centering 0.150 & \centering 0.537 & \centering 0.498 & \centering 0.515 & \centering\arraybackslash 0.466\\
& ITA-ECBS-v0 & \centering 0.864 & \centering 0.589 & \centering 0.379 & \centering 0.190 & \centering 0.150  & \centering 0.545 & \centering 0.498 & \centering 0.487 & \centering\arraybackslash 0.427 \\
& ITA-ECBS & \centering \textbf{0.875} & \centering \textbf{0.639} & \centering \textbf{0.453} & \centering \textbf{0.267} & \centering \textbf{0.223} & \centering \textbf{0.578} & \centering \textbf{0.562} & \centering \textbf{0.564} & \centering\arraybackslash \textbf{0.480} \\ \hline
\multirow{3}{*}{\textbf{1.03}} & ECBS\_TA & \centering 0.873 & \centering 0.782 & \centering 0.640 & \centering 0.406 & \centering 0.200   & \centering 0.698 & \centering 0.610 & \centering 0.625 & \centering\arraybackslash 0.600\\
& ITA-ECBS-v0 & \centering 0.878 & \centering 0.762 & \centering 0.603 & \centering 0.376 & \centering 0.215   & \centering 0.701 & \centering 0.635 & \centering 0.618 & \centering\arraybackslash 0.528\\
& ITA-ECBS & \centering \textbf{0.890} & \centering \textbf{0.820} & \centering \textbf{0.707} & \centering \textbf{0.515} & \centering \textbf{0.387} & \centering \textbf{0.76} & \centering \textbf{0.741} & \centering \textbf{0.719} & \centering\arraybackslash \textbf{0.612}\\ \hline
\multirow{3}{*}{\textbf{1.04}} & ECBS\_TA & \centering 0.889 & \centering 0.817 & \centering 0.701 & \centering 0.482 & \centering 0.279   & \centering 0.743 & \centering 0.647 & \centering 0.673 & \centering\arraybackslash 0.660\\
& ITA-ECBS-v0 & \centering 0.893 & \centering 0.792 & \centering 0.657 & \centering 0.471 & \centering 0.315   & \centering 0.747 & \centering 0.678 & \centering 0.672 & \centering\arraybackslash 0.582\\
& ITA-ECBS & \centering \textbf{0.906} & \centering \textbf{0.843} & \centering \textbf{0.770} & \centering \textbf{0.607} & \centering \textbf{0.490} & \centering \textbf{0.795} & \centering \textbf{0.787} & \centering \textbf{0.777} & \centering\arraybackslash \textbf{0.672}\\ \hline
\multirow{3}{*}{\textbf{1.05}} & ECBS\_TA & \centering 0.898 & \centering 0.837 & \centering 0.737 & \centering 0.529 & \centering 0.320   & \centering 0.771 & \centering 0.669 & \centering 0.703 & \centering\arraybackslash 0.701\\
& ITA-ECBS-v0 & \centering 0.904 & \centering 0.810 & \centering 0.696 & \centering 0.509 & \centering 0.373   & \centering 0.775 & \centering 0.719 & \centering 0.702 & \centering\arraybackslash 0.607\\
& ITA-ECBS & \centering \textbf{0.932} & \centering \textbf{0.859} & \centering \textbf{0.810} & \centering \textbf{0.671} & \centering \textbf{0.551} & \centering \textbf{0.831} & \centering \textbf{0.822} & \centering \textbf{0.814} & \centering\arraybackslash \textbf{0.708}\\ \hline
\multirow{3}{*}{\textbf{1.10}} & ECBS\_TA & \centering 0.959 & \centering 0.868 & \centering 0.85 & \centering 0.692 & \centering 0.485   & \centering 0.871 & \centering 0.752 & \centering 0.787 & \centering\arraybackslash \textbf{0.820}\\
& ITA-ECBS-v0 & \centering 0.964 & \centering 0.848 & \centering 0.804 & \centering 0.684 & \centering 0.571   & \centering 0.868 & \centering 0.842 & \centering 0.806 & \centering\arraybackslash 0.699\\
& ITA-ECBS & \centering \textbf{0.985} & \centering \textbf{0.873} & \centering \textbf{0.854} & \centering \textbf{0.801} & \centering \textbf{0.757} & \centering \textbf{0.901} & \centering \textbf{0.902} & \centering \textbf{0.889} & \centering\arraybackslash 0.790\\ \hline
\end{tabular}
\caption{Average success rates across variables not specified in the table. The highest average success rates are shown in bold. Due to space constraints, only five numbers of agents are shown.}
\label{tab:my-table}
\end{table*}

\Cref{alg:ITA-ECBS1} shows the pseudocode of ITA-ECBS-v0. It starts by generating the root node $H_0$, including creating an empty constraint set \(H_0.\Omega\) and the corresponding matrices $H_0.M_c$ and $H_0.M_L$ using focal search.
Subsequently, a target assignment algorithm, such as the dynamic Hungarian algorithm, determines $H_0.\pi_{ta}$ based on $H_0.M_L$. Then $H_0.\pi$ is obtained from $H_0.M_c$ based on $H_0.\pi_{ta}$ (Lines 1-9).
In the high-level search, ITA-ECBS-v0, like ECBS, maintains two priority queues: OPEN and FOCAL. OPEN has all CT nodes sorted by their $c_L$. FOCAL contains only those CT nodes $H$ with $H.c \leq w \cdot H_{front}.c_L$, where $H_{front}$ is the OPEN front node. FOCAL is sorted by a given heuristic function $d(H)$. ITA-ECBS-v0 first gets \(H_{front}\) from OPEN in each iteration. Based on \(H_{front}.c_L\), it adds all eligible CT nodes in OPEN to FOCAL. Then, it chooses the front node \(H_{cur}\) in FOCAL and removes it from OPEN (Lines 10-16).

ITA-ECBS-v0 checks if $H_{cur}.\pi$ is collision-free. If so, this $\pi$ is a bounded-suboptimal valid solution (Lines 17-18).
Otherwise, like ECBS, ITA-ECBS-v0 utilizes the first identified collision to generate two constraints.
It then creates two child CT nodes identical to \(H_{cur}\) and adds one constraint to \(\Omega\) of the first child CT node and the other constraint to \(\Omega\) of the second child CT node (Lines 19-25).
For each child CT node \(Q\) with a constraint for agent $k$ added, all LB values in $Q.M_L$ and all paths in $Q.M_c$ related to agent $k$ have to be replanned subject to the new constraint set $Q.\Omega$ (Lines 26-27, 36).
Since \(Q.M_L\) changes, the TA $Q_{\pi_{ta}}$, solution $Q.\pi$, and costs $Q.c$ and $Q.c_L$ have to be updated as well (Lines 28-30). Because focal search returns $\infty$ if no solution exists, $Q.c = \infty$ means that there is no \(\pi\) satisfying $Q.\Omega$ and ITA-ECBS-v0 can ignore this CT node. Otherwise, it insert $Q$ into OPEN (Lines 31-32).

Usually, bounded-suboptimal algorithms aim to find a bounded-suboptimal valid solution swiftly. In the low-level search, we could have more candidates in FOCAL by obtaining larger LB values. Since all candidates in FOCAL have a bounded-suboptimal solution, more candidates in FOCAL increases our chances of rapidly discovering a valid solution if $d(n)$ is properly designed.
For an LB value $c^g$, we have $c^g \leq c^{opt}$ where $c^{opt}$ is the shortest path cost. Rather than obtaining an LB value $c^g$ from focal search, we can directly use $c^{opt}$ as the LB value from a shortest path algorithm such as $A^*$. 
Using $c^{opt}$ as the LB value can give us more freedom to get a path leading to a valid \(\pi\).
We use a new function named $searchWithLB$ to identify paths with the lowest $d(n)$ values, using an given LB value $c^v$ as a guideline. $searchWithLB$ is similar to focal search but only has FOCAL contains all candidates with costs  $w\cdot c^{v}$. If a candidate cost is larger than $w\cdot c^{v}$, it cannot contain a bounded-suboptimal solution and we can ignore it.

The final version of ITA-ECBS is shown in \Cref{alg:ITA-ECBS1}. The only difference between ITA-ECBS-v0 and ITA-CBS lies in Lines 37-39. 
We first use shortest path search to determine $c^{opt}$ based on \(\Omega\). $c^{opt}$ is then utilized as $c^v$ in the $searchWithLB$ function to identify a path with a cost of at most $w\cdot c^{opt}$, thus enabling quicker discovery of a bounded-suboptimal valid solution.
As demonstrated in our experimental section, ITA-ECBS is more efficient than ITA-ECBS-v0 despite requiring twice path searches. 

In code implementation, we use the number of collisions in $H.\pi$ as the heuristic function \(d\) for both the low-level path search and the high-level CT node search. However, this heuristic is more advantageous for the low-level search of ECBS-TA than ITA-ECBS. ECBS-TA updates a single path for one agent and does not need to modify the TA solution of CT nodes, which ensures \(d(n)\) is accurate when using low-level focal search to find a path. In ITA-ECBS, the accuracy is compromised because changes in the TA can lead to changes in the paths of multiple agents. When calculating the number of collisions for a path in ITA-ECBS, the paths of the other agents are not finalized. For example, we need to re-plan paths for two agents $a_1$ and $a_2$ one by one due to a new \(\pi_{ta}\). We want to calculate the number of collisions for the focal search of $a_1$. However, it is impossible to accurately count collisions between $a_1$ and $a_2$ because we do not know $a_2$'s new path. So we can only use $a_2$'s old path to count collisions between $a_1$ and $a_2$.

\section{Experimental Results}

We compare the success rate and runtime of ITA-ECBS with ECBS-TA since, to the best of our knowledge, ECBS-TA is the only bounded-suboptimal TAPF algorithm that minimizes flowtime.
We implement ITA-ECBS and ECBS-TA in C++, partially based on the existing ITA-CBS implementation.\footnote{
The ITA-ECBS and ECBS-TA code and test data are available at \url{https://github.com/TachikakaMin/ITA-CBS2}.
Based on tests, our ECBS-TA implementation runs faster than the original one.} 
All experiments are conducted on an Ubuntu 20.04.1 system with an AMD Ryzen 3990X 64-Core Processor with 2133 MHz 64GB RAM.

\subsection{Test Settings} 

We test ITA-ECBS and ECBS-TA with the suboptimality factors \(w=1.00, 1.01, 1.02, 1.03, 1.04, 1.05, 1.10\), and \(1.20\) on 8 maps from the MAPF Benchmark~\cite{stern2019mapf}. These maps are shown in \Cref{fig:w_cr} and previously to evaluate ITA-CBS and CBS-TA~\cite{tang2023solving}: (1) random-32-32-10 (32x32) and empty-32-32 (32x32) are grid maps with and without random obstacles, (2) den312d (65x81) is a grid map from the video game Dragon Age Origins, (3) maze-32-32-2 (32x32) is a maze-like grid map, (4) room-64-64-8 (64x64) is a room-like grid map, (5) warehouse-10-20-10-2-1 (161x63) is a grid map inspired by real-world autonomous warehouses, and (6) orz900d (1491x656) and Boston-0-256 (256x256) are the largest and second largest benchmark maps.  

The number of agents ranges from 10 to 150 in increments of 10. For each map, every agent has a target set of the same size, which is 5, 15, 5, 4, 20, 30, 10, 10 for maps random-32-32-10, den312d, empty-32-32, maze-32-32-2, room-64-64-8, warehouse-10-20-10-2-1, orz900d, Boston-0-256 respectively.
Each target set has unique targets and targets shared among all agents. We vary the percentage of shared targets in target sets from 0\%, 30\%, 60\% to 100\%. All numbers round down.%
\footnote{The size of the target set for each map is determined by the number of cells available to be assigned to agents, except large maps orz900d and Boston-0-256. For instance, the empty-32-32 map has 1,024 cells. With a maximum of 150 agents, the target set size is calculated as \(1,024 / 150 = 6.82\), and we use 5 as the target set size. On large maps, the size of the target set is 10 to prevent ITA-ECBS and ECBS-TA from timing out. Because most of the time on a large map will be occupied by low-level searches, the number of CT nodes that can be searched within 30 seconds is only a few dozen if the size of the target set is larger.} 
However, we ensure that each target set always includes at least one unique target to guarantee agents have enough targets to allocate.

For each map, number of agents, and percentage of shared targets, we generate 20 test cases with randomly selected start and target locations. An algorithm is considered to have failed for a given test case if it does not find a valid solution within 30 seconds. The success rate is the percentage of the 20 test cases where the algorithm succeeds. 

\begin{figure}[t!]
\centering
\includegraphics[width=0.45\textwidth]{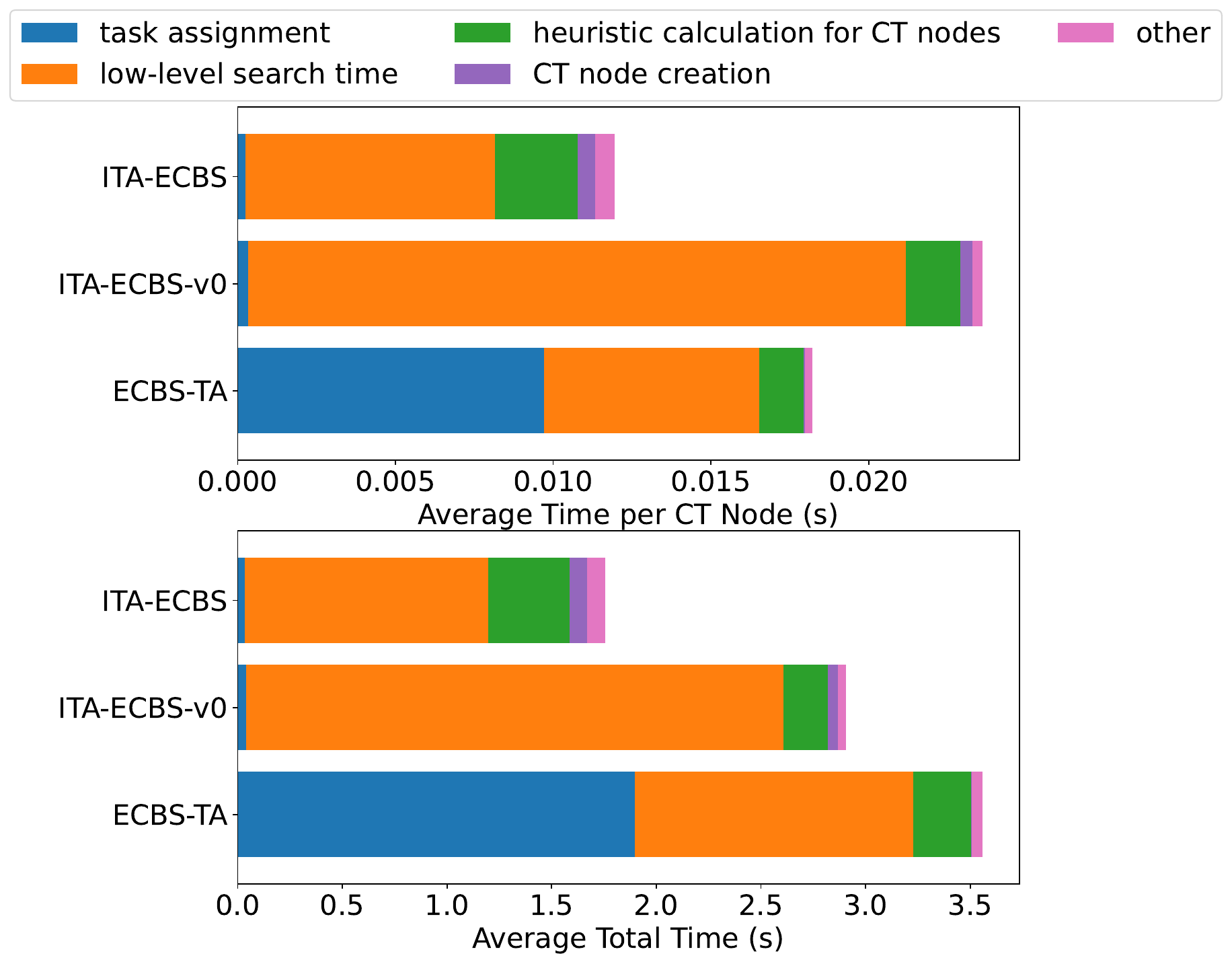}
\caption{Runtime breakdown (seconds) for
target assignment (\Cref{alg:ITA-ECBS1} Line 28),
low-level search (Line 27),
heuristic calculation for CT nodes (Line 32), 
CT node creation (Line 21), which requires copying variables and other tasks. 
}
\label{fig:test3}
\end{figure}

\subsection{Performance}
Overall, we have 76,800 test cases. Among these, 48,334 test cases are solved by both ITA-ECBS and ECBS-TA, 5,190 are solved only by ITA-ECBS, 509 are solved only by ECBS-TA, and 22,767 are not solved by either algorithm. Out of the 54,033 test cases solved by at least one algorithm, ITA-ECBS was faster than ECBS-TA in 87.42\% of the test cases and 5 times faster in 24.71\% of them. \Cref{fig:g3} showcases their performance for three selected suboptimality factors. As the suboptimality factor increases, ITA-ECBS and ECBS-TA solve more test cases and the success rates of ITA-ECBS are larger than those of ECBS-TA.

\Cref{fig:w_cr} displays the success rates of different algorithms as the suboptimality factor increases. \Cref{fig:w_anum_cr} displays the success rates of different algorithms as the number of agents increases for three selected suboptimality factors. The success rates of all algorithms tend to decrease as the number of agents increases, as expected. The success rates of all algorithms increase as the suboptimality factor increases. In orz900d, the success rate of ITA-ECBS-v0 decreases significantly, likely because orz900d is the largest map (1491x656) and the low-level focal search of ITA-ECBS-v0 is slow due to the uninformed heuristic function. \Cref{tab:my-table} summarizes the average success rates for various algorithms as a function of their suboptimality factors, numbers of agents, and percentages of shared targets. ITA-ECBS outperforms ECBS-TA in most scenarios.


\Cref{fig:test3} shows the average runtimes of different components of three algorithms, on those test cases that they solved within the runtime limit across different suboptimality factors.
ITA-ECBS-v0 is slower than ECBS-TA in processing each CT node, primarily due to its slower low-level focal search. But it is faster than ECBS-TA because ITA-ECBS-v0 uses a single CT and generates fewer CT nodes. ITA-ECBS improves on ITA-ECBS-v0 by obtaining a larger LB value to increase the number of nodes in FOCAL, which reduces the path search time both per CT node and the average time. Additionally, ITA-ECBS and ITA-ECBS-v0 benefit from using the dynamic Hungarian algorithm, which is considerably faster than the next-best target assignment algorithm used in ECBS-TA, as their task-assignment runtimes show.

\section{Conclusion}

This work presented a new algorithm, Incremental Target Assignment with Enhanced CBS (ITA-ECBS), designed to solve the TAPF problem with a bounded-suboptimal flowtime. It is the first bounded-suboptimal algorithm derived from ITA-CBS, a leading optimal algorithm for TAPF. By using an LB matrix to derive the TA solution, ITA-ECBS avoids the unbounded problem, a risk present when directly converting the CBS algorithm of ITA-CBS to its bounded-suboptimal version ECBS. Furthermore, ITA-ECBS uses shortest path costs as LB values, which accelerate the focal search for pathfinding. Although ITA-ECBS could be improved further, such as by designing a good heuristic function for its CT nodes despite different CT nodes having different TA solutions, our experimental results demonstrate that ITA-ECBS is significantly faster than the prior best bounded-suboptimal TAPF algorithm ECBS-TA.

\section{Acknowledgments}
Our research was supported by the National Science Foundation (NSF) under grant numbers 1817189, 1837779, 1935712, 2121028, 2112533, and 2321786 as well as gifts from Amazon Robotics and the CMU Manufacturing Futures Institute, made possible by the Richard King Mellon Foundation. The views and conclusions contained in this document are those of the authors and should not be interpreted as representing the official policies, either expressed or implied, of the sponsoring organizations, agencies, or the U.S. government.

\bibliography{strings,aaai24}

\begin{thebibliography}{31}
\providecommand{\natexlab}[1]{#1}

\bibitem[{Barer et~al.(2014)Barer, Sharon, Stern, and
  Felner}]{barer2014suboptimal}
Barer, M.; Sharon, G.; Stern, R.; and Felner, A. 2014.
\newblock Suboptimal variants of the conflict-based search algorithm for the
  multi-agent pathfinding problem.
\newblock In \emph{{Proceedings of the International Symposium on Combinatorial
  Search (SoCS)}}, volume~5, 19--27.

\bibitem[{Boyarski et~al.(2015)Boyarski, Felner, Stern, Sharon, Betzalel,
  Tolpin, and Shimony}]{boyarski2015icbs}
Boyarski, E.; Felner, A.; Stern, R.; Sharon, G.; Betzalel, O.; Tolpin, D.; and
  Shimony, E. 2015.
\newblock {ICBS}: The improved conflict-based search algorithm for multi-agent
  pathfinding.
\newblock In \emph{{Proceedings of the International Symposium on Combinatorial
  Search (SoCS)}}, volume~6, 223--225.

\bibitem[{Chan et~al.(2023)Chan, Stern, Felner, and Koenig}]{chan2023greedy}
Chan, S.-H.; Stern, R.; Felner, A.; and Koenig, S. 2023.
\newblock Greedy priority-based search for suboptimal multi-agent path finding.
\newblock In \emph{{Proceedings of the International Symposium on Combinatorial
  Search (SoCS)}}, volume~16, 11--19.

\bibitem[{Chegireddy and Hamacher(1987)}]{chegireddy1987algorithms}
Chegireddy, C.~R.; and Hamacher, H.~W. 1987.
\newblock {Algorithms for finding k-best perfect matchings}.
\newblock \emph{Discrete Applied Mathematics}, 18(2): 155--165.

\bibitem[{Chen et~al.(2021)Chen, Alonso-Mora, Bai, Harabor, and
  Stuckey}]{chen2021integrated}
Chen, Z.; Alonso-Mora, J.; Bai, X.; Harabor, D.; and Stuckey, P.~J. 2021.
\newblock {Integrated task assignment and path planning for capacitated
  multi-agent pickup and delivery}.
\newblock \emph{{IEEE Robotics and Automation Letters}}, 6(3): 5816--5823.

\bibitem[{Cohen et~al.(2018)Cohen, Greco, Ma, Hern{\'a}ndez, Felner, Kumar, and
  Koenig}]{cohen2018anytime}
Cohen, L.; Greco, M.; Ma, H.; Hern{\'a}ndez, C.; Felner, A.; Kumar, T.~S.; and
  Koenig, S. 2018.
\newblock Anytime focal search with applications.
\newblock In \emph{{Proceedings of the International Joint Conference on
  Artificial Intelligence (IJCAI)}}, 1434--1441.

\bibitem[{Engquist(1982)}]{engquist1982successive}
Engquist, M. 1982.
\newblock A successive shortest path algorithm for the assignment problem.
\newblock \emph{Information Systems and Operational Research (INFOR)}, 20(4):
  370--384.

\bibitem[{Erdmann and Lozano-Perez(1987)}]{erdmann1987multiple}
Erdmann, M.; and Lozano-Perez, T. 1987.
\newblock On multiple moving objects.
\newblock \emph{Algorithmica}, 2: 477--521.

\bibitem[{H\"{o}nig et~al.(2018)H\"{o}nig, Kiesel, Tinka, Durham, and
  Ayanian}]{honig2018conflict}
H\"{o}nig, W.; Kiesel, S.; Tinka, A.; Durham, J.~W.; and Ayanian, N. 2018.
\newblock Conflict-based search with optimal task assignment.
\newblock In \emph{{Proceedings of the International Conference on Autonomous
  Agents and Multiagent Systems (AAMAS)}}, 757--765.

\bibitem[{Li et~al.(2022)Li, Chen, Harabor, Stuckey, and Koenig}]{li2022mapf}
Li, J.; Chen, Z.; Harabor, D.; Stuckey, P.~J.; and Koenig, S. 2022.
\newblock MAPF-LNS2: Fast repairing for multi-agent path finding via large
  neighborhood search.
\newblock In \emph{{Proceedings of the AAAI Conference on Artificial
  Intelligence (AAAI)}}, volume~36, 10256--10265.

\bibitem[{Li, Ruml, and Koenig(2021)}]{li2021eecbs}
Li, J.; Ruml, W.; and Koenig, S. 2021.
\newblock EECBS: A bounded-suboptimal search for multi-agent path finding.
\newblock In \emph{{Proceedings of the AAAI Conference on Artificial
  Intelligence (AAAI)}}, volume~35, 12353--12362.

\bibitem[{Luna and Bekris(2011)}]{luna2011push}
Luna, R.~J.; and Bekris, K.~E. 2011.
\newblock Push and swap: Fast cooperative path-finding with completeness
  guarantees.
\newblock In \emph{{Proceedings of the International Joint Conference on
  Artificial Intelligence (IJCAI)}}, 294--300.

\bibitem[{Ma et~al.(2019)Ma, Harabor, Stuckey, Li, and
  Koenig}]{ma2019searching}
Ma, H.; Harabor, D.; Stuckey, P.~J.; Li, J.; and Koenig, S. 2019.
\newblock Searching with consistent prioritization for multi-agent path
  finding.
\newblock In \emph{{Proceedings of the AAAI Conference on Artificial
  Intelligence (AAAI)}}, volume~33, 7643--7650.

\bibitem[{Ma and Koenig(2016)}]{cbm2016}
Ma, H.; and Koenig, S. 2016.
\newblock Optimal target assignment and path finding for teams of agents.
\newblock In \emph{{Proceedings of the International Conference on Autonomous
  Agents and Multiagent Systems (AAMAS)}}, 1144--1152.

\bibitem[{Mills-Tettey, Stentz, and Dias(2007)}]{mills2007dynamic}
Mills-Tettey, G.~A.; Stentz, A.; and Dias, M.~B. 2007.
\newblock {The dynamic {Hungarian} algorithm for the assignment problem with
  changing costs}.
\newblock \emph{Robotics Institute, Pittsburgh, PA, Tech. Rep.
  CMU-RI-TR-07-27}.

\bibitem[{Okumura(2023)}]{okumura2023lacam}
Okumura, K. 2023.
\newblock LaCAM: Search-based algorithm for quick multi-agent pathfinding.
\newblock In \emph{{Proceedings of the AAAI Conference on Artificial
  Intelligence (AAAI)}}, volume~37, 11655--11662.

\bibitem[{Okumura and Défago(2023)}]{okumura2023solving}
Okumura, K.; and Défago, X. 2023.
\newblock Solving simultaneous target assignment and path planning efficiently
  with time-independent execution.
\newblock \emph{Artificial Intelligence}, 321: 103946.

\bibitem[{Pearl and Kim(1982)}]{pearl1982studies}
Pearl, J.; and Kim, J.~H. 1982.
\newblock Studies in semi-admissible heuristics.
\newblock \emph{{IEEE} Transactions on Pattern Analysis and Machine
  Intelligence (PAMI)}, 4: 392--399.

\bibitem[{Ren, Rathinam, and Choset(2023)}]{ren23cbss}
Ren, Z.; Rathinam, S.; and Choset, H. 2023.
\newblock {CBSS}: A new approach for multiagent combinatorial path finding.
\newblock \emph{IEEE Transactions on Robotics}, 39(4): 2669--2683.

\bibitem[{Sharon et~al.(2015)Sharon, Stern, Felner, and
  Sturtevant}]{sharon2015conflict}
Sharon, G.; Stern, R.; Felner, A.; and Sturtevant, N.~R. 2015.
\newblock {Conflict-based search for optimal multi-agent pathfinding}.
\newblock \emph{Artificial Intelligence}, 219: 40--66.

\bibitem[{Silver(2005)}]{silver2005cooperative}
Silver, D. 2005.
\newblock Cooperative pathfinding.
\newblock In \emph{{Proceedings of the AAAI Conference on Artificial
  Intelligence and Interactive Digital Entertainment (AIIDE)}}, volume~1,
  117--122.

\bibitem[{Standley(2010)}]{standley2010finding}
Standley, T. 2010.
\newblock {Finding optimal solutions to cooperative pathfinding problems}.
\newblock In \emph{{Proceedings of the AAAI Conference on Artificial
  Intelligence (AAAI)}}, volume~24, 173--178.

\bibitem[{Standley and Korf(2011)}]{standley2011complete}
Standley, T.; and Korf, R. 2011.
\newblock Complete algorithms for cooperative pathfinding problems.
\newblock In \emph{{Proceedings of the International Joint Conference on
  Artificial Intelligence (IJCAI)}}, 668--673.

\bibitem[{Stern et~al.(2019)Stern, Sturtevant, Felner, Koenig, Ma, Walker, Li,
  Atzmon, Cohen, Kumar, Boyarski, and Bartak}]{stern2019mapf}
Stern, R.; Sturtevant, N.~R.; Felner, A.; Koenig, S.; Ma, H.; Walker, T.~T.;
  Li, J.; Atzmon, D.; Cohen, L.; Kumar, T. K.~S.; Boyarski, E.; and Bartak, R.
  2019.
\newblock {Multi-Agent pathfinding: Definitions, variants, and benchmarks}.
\newblock In \emph{{Proceedings of the International Symposium on Combinatorial
  Search (SoCS)}}, 1, 151--158.

\bibitem[{Tang et~al.(2023)Tang, Ren, Li, and Sycara}]{tang2023solving}
Tang, Y.; Ren, Z.; Li, J.; and Sycara, K. 2023.
\newblock Solving multi-agent target assignment and path finding with a single
  constraint tree.
\newblock In \emph{{IEEE} International Symposium on Multi-Robot and
  Multi-Agent Systems (MRS)}, 8--14.

\bibitem[{Wagner and Choset(2011)}]{wagner2011m}
Wagner, G.; and Choset, H. 2011.
\newblock M*: A complete multirobot path planning algorithm with performance
  bounds.
\newblock In \emph{{Proceedings of the {IEEE/RSJ} International Conference on
  Intelligent Robots and Systems (IROS)}}, 3260--3267.

\bibitem[{Wagner and Choset(2015)}]{wagner2015subdimensional}
Wagner, G.; and Choset, H. 2015.
\newblock {Subdimensional expansion for multirobot path planning}.
\newblock \emph{Artificial Intelligence}, 219: 1--24.

\bibitem[{Wang and Botea(2008)}]{wang2008fast}
Wang, K.-H.~C.; and Botea, A. 2008.
\newblock Fast and memory-efficient multi-agent pathfinding.
\newblock In \emph{{Proceedings of the International Conference on Automated
  Planning and Scheduling (ICAPS)}}, 380--387.

\bibitem[{Wurman, D'Andrea, and Mountz(2008)}]{wurman2008coordinating}
Wurman, P.~R.; D'Andrea, R.; and Mountz, M. 2008.
\newblock Coordinating hundreds of cooperative, autonomous vehicles in
  warehouses.
\newblock \emph{AI Magazine}, 29(1): 9--19.

\bibitem[{Yu and LaValle(2013)}]{yu2013structure}
Yu, J.; and LaValle, S. 2013.
\newblock {Structure and intractability of optimal multi-robot path planning on
  graphs}.
\newblock In \emph{{Proceedings of the AAAI Conference on Artificial
  Intelligence (AAAI)}}, volume~27, 1443--1449.

\bibitem[{Zhong et~al.(2022)Zhong, Li, Koenig, and Ma}]{zhong2022optimal}
Zhong, X.; Li, J.; Koenig, S.; and Ma, H. 2022.
\newblock Optimal and bounded-suboptimal multi-goal task assignment and path
  finding.
\newblock In \emph{{Proceedings of IEEE International Conference on Robotics
  and Automation (ICRA)}}, 10731--10737.

\end{thebibliography}

\end{document}